\documentclass[conference]{IEEETran}
\IEEEoverridecommandlockouts

\usepackage{cite}
\usepackage{amsmath,amssymb,amsfonts}
\usepackage{algorithmic}
\usepackage{graphicx}
\usepackage{textcomp}
\usepackage{xcolor}

\usepackage{multirow}
\usepackage{amsmath}
\usepackage{amssymb}
\usepackage{makecell}
\usepackage{bm}
\usepackage{gensymb}
\usepackage{booktabs}       

\def\BibTeX{{\rm B\kern-.05em{\sc i\kern-.025em b}\kern-.08em
    T\kern-.1667em\lower.7ex\hbox{E}\kern-.125emX}}
\begin{document}

\title{Realistic and Controllable 3D Gaussian-Guided Object Editing for Driving Video Generation}

\author{Jiusi Li$^{1}$, Jackson Jiang$^{2}$, Jinyu Miao$^{1}$, Miao Long$^{1}$, Tuopu Wen$^{1}$, Peijin Jia$^{1}$, Shengxiang Liu$^{2}$ \\ Chunlei Yu$^{2}$, Maolin Liu$^{2}$, Yuzhan Cai$^{3}$, Kun Jiang$^{1}$$^*$, Mengmeng Yang$^{1}$, Diange Yang$^{1}$$^*$

\thanks{$^{1}$Jiusi Li, Jinyu Miao, Miao Long, Tuopu Wen, Peijin Jia, Kun Jiang, Mengmeng Yang and Diange Yang are with
the School of Vehicle and Mobility, and State Key Laboratory of Intelligent Green Vehicle and Mobility, Tsinghua University, Beijing, 100084, China. (li-js23@mails.tsinghua.edu.cn.)        }%
\thanks{$^{2}$Jackson Jiang, Shengxiang Liu, Chunlei Yu and Maolin Liu are with WUWEN AI, Beijing 100084, China.
}
\thanks{$^{3}$Yuzhan Cai is with PhiGent Robotics, Beijing 100084, China.
}
\thanks{$^*$Corresponding author.
}
}

\maketitle
\begin{abstract}

Corner cases are crucial for training and validating autonomous driving systems, yet collecting them from the real world is often costly and hazardous. Editing objects within captured sensor data offers an effective alternative for generating diverse scenarios, commonly achieved through 3D Gaussian Splatting or image generative models. However, these approaches often suffer from limited visual fidelity or imprecise pose control.
To address these issues, we propose G\textsuperscript{2}Editor, a framework designed for photorealistic and precise object editing in driving videos. Our method leverages a 3D Gaussian representation of the edited object as a dense prior, injected into the denoising process to ensure accurate pose control and spatial consistency. A scene-level 3D bounding box layout is employed to reconstruct occluded areas of non-target objects. Furthermore, to guide the appearance details of the edited object, we incorporate hierarchical fine-grained features as additional conditions during generation.
Experiments on the Waymo Open Dataset demonstrate that G\textsuperscript{2}Editor effectively supports object repositioning, insertion, and deletion within a unified framework, outperforming existing methods in both pose controllability and visual quality, while also benefiting downstream data-driven tasks.
 
\end{abstract}


\section{Introduction}\label{sec:intro} 

Extensive data serves as a foundation for advanced Autonomous Driving (AD) systems. 
In recent years, large-scale public datasets have played a crucial role in AD model training and validation.
However, real-world data often exhibits a long-tail distribution \cite{survey, synthetic}, making it difficult for models to handle rare yet critical corner cases, which are costly and risky to collect. 
This challenge has spurred increasing interest in synthetic data generation to supplement real-world data and enhance the robustness and safety of AD systems. 

Although existing data generation methods can generate various driving scenarios, they struggle to produce high-fidelity synthetic data with precise pose control. Some works leverage diffusion models to generate full-scene data conditioned on scene layouts (e.g., 3D bounding boxes, maps) and text, producing semantically aligned visual data from noise \cite{drivescape,panacea}. Yet, these approaches struggle to enable fine-grained control over individual object poses and appearances, failing to support directional generation grounded in real-world data. To address this, some methods focus on editing objects within captured data \cite{GenMM,DriveEditor}. They often frame the task as diffusion-based image inpainting, using 3D bounding boxes to control object poses. While diffusion priors can guide appearance via reference images, 3D box-based pose signals provide insufficient constraints for drastic pose changes, leading to inaccurate object geometry. 

To ensure accurate geometric control for edited objects in driving scenes editing, some recent methods instead utilize neural rendering techniques, such as Neural Radiance Fields (NeRF) and 3D Gaussian Splatting (3DGS). Some methods like \cite{drivinggaussian, OmniRe} decompose driving scenes into background and foreground objects, modeling objects as individual 3DGS instances to enable accurate geometric control. However, such solution are prone to degradation due to limited and sparse vehicle-mounted views, causing rendering distortion under large pose variations. Additionally, the lack of explicit lighting modeling introduces fidelity issues, such as inconsistent illumination between edited objects and backgrounds.  

\begin{figure*}[t]
    \centering
    \includegraphics[width=1.0\linewidth]{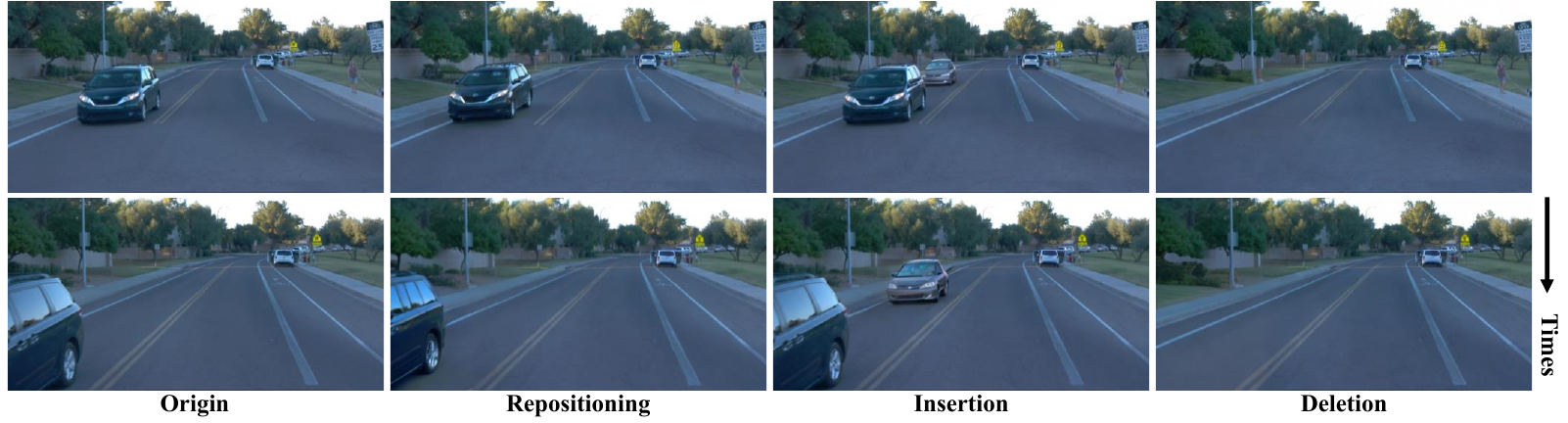}
    \caption{G\textsuperscript{2}Editor enables realistic and controllable object editing, including repositioning, insertion and deletion.}
    \label{fig:intro}
\end{figure*}

To bridge precise control and high-fidelity synthesis, we introduce G\textsuperscript{2}Editor, a framework for realistic and geometrically accurate object editing in driving scenes. By integrating the photorealistic appearance synthesis of diffusion models with the explicit geometric control of 3DGS, G\textsuperscript{2}Editor achieves dual objectives: 
1) \textit{Precise Pose Control}: A geometric-aware control signal is proposed that combines projected 3D bounding boxes and rendered 3D Gaussian models onto the image plane, providing dense and accurate spatial positional cues to guide the diffusion denoising process.  
2) \textit{Appearance Fidelity and Temporal Consistency}: Appearance details are captured from the reference image, which are hierarchically injected into the denoising UNet via spatial attention. Temporal layers are trained separately to ensure content consistency across frames.  
Experiments on Waymo Open Dataset \cite{waymo} demonstrate that G\textsuperscript{2}Editor enables photorealistic object repositioning, insertion, and deletion, as shown in  Fig.~\ref{fig:intro}, outperforming state-of-the-art methods in pose control accuracy and visual fidelity. Our contributions can be summarized as follows: 
  
\begin{itemize}  
 
\item We propose G\textsuperscript{2}Editor, an framework for driving video editing that supports realistic object manipulation.  
 
\item We introduce a hybrid pose control strategy combining 3D boxes and 3D Gaussian, ensuring precise control and spatial coherence during editing. A scene-level 3D bounding box layout is employed to effectively infer and recover occluded parts of non-target objects.  

\item Experiments on the public dataset demonstrate the state-of-the-art performance of G\textsuperscript{2}Editor in terms of pose accuracy and visual fidelity. The generated data can boost downstream AD tasks by providing high-quality and diverse driving scenes.
\end{itemize}

\section{Related Works}\label{sec:related_works}
\textit{Full-scene Driving Video Generation:}
To tackle the diversity limitations of real world driving data, researchers have employed generative techniques to expand the variety of driving videos. Many efforts focus on controllable full-scene generation that synthesizes street-view images \cite{bevgen, critical} or driving videos \cite{drivescape,panacea} aligned with semantic driving scenarios represented via 3D layouts and text prompts. However, these semantic-level controls lack fine-grained object-level constraints, thereby limiting the ability to precisely manipulate individual objects. To enhance object-level manipulation, SubjectDrive \cite{subjectdrive} introduces a subject bank mechanism. Despite the progress in scalable AD data generation, these methods still struggle to achieve detailed control over specific objects, failing to support directional editing grounded in real-world data for high-fidelity synthesis of long-tail scenarios.


\textit{Diffusion Models for Image and Video Editing:}
Some recent works have shifted the focus from full-scene visual data generation to object-level editing in driving videos. 
With the rapid development of Stable Diffusion (SD) \cite{sd}, many diffusion-based models support image editing driven by text \cite{emuedit,PowerPaint} or reference images \cite{objectstitch, PbE}. Although these methods achieve seamless image editing at the 2D level, they lack precise control over object pose and spatial consistency in 3D space, limiting their applicability in AD scenarios.
Recently, inspired by these works, GenMM \cite{GenMM} adopts the inpainting framework to tackle driving video editing, enabling object insertion based on a single reference image. DriveEditor \cite{DriveEditor} further introduces a unified framework supporting multiple editing tasks. However, existing methods still face challenges in accurate pose control and appearance maintenance. To this end, we explicitly incorporate the 3D Gaussian to enhance pose accuracy and improve spatial coherence.

\textit{NeRF and 3D Gaussian Editing:}
With the development of NeRF \cite{NeRF} and 3DGS \cite{3DGS}, numerous NeRF-based \cite{mars,unisim} and 3DGS-based \cite{drivinggaussian,OmniRe} methods have been developed to separately represent dynamic foregrounds and static backgrounds, enabling object editing and simulation in driving scenarios. OmniRe \cite{OmniRe}, for instance, decomposes static backgrounds, vehicles, and dynamic actors into separate nodes to facilitate object editing. While these methods enable precise object pose control, they face challenges in flexible object editing due to inherent overfitting to training views, along with fidelity issues such as inconsistent lighting and poor shadow rendering.
Other approaches like Lift3D \cite{li2023lift3d} and GINA-3D \cite{shen2023gina} use image synthesis networks or NeRF to construct full 3D assets and integrate them into driving scenes. These approaches also suffer from visual artifacts, most notably in inconsistent illumination and misaligned shadows at object boundaries. Our method incorporates 3D Gaussian information into the diffusion-based framework to enhance physical and visual realism.
\section{Method}\label{sec:method}

\subsection{Problem Formulation: Object Editing as Video Editing}
\label{sec:preli}
{
\setlength{\parindent}{0cm}
\textbf{Preliminaries: Stable Diffusion \cite{sd}.}
Our framework is based on SD, which encodes images in a latent space. A denoising UNet learns to predict the added noise $\epsilon$ to the image latent $\mathbf{z}$. This optimization objective can be formulated as:
}
\begin{equation}
    \mathcal{L} = \mathbb{E}_{\mathbf{z}_t,\mathbf{c}_c,\epsilon,t}[\left \| \epsilon -  \epsilon_{\theta}(\mathbf{z}_t,\mathbf{c}_c,t) \right \| ^{2}_2],
\end{equation}
where $\mathbf{z}_t$ is the noisy latent at step $t$ and $\mathbf{c}_c$ represents conditional embeddings. During inference, a latent $\mathbf{z}_T$ is sampled from Gaussian noise and progressively denoised to $\mathbf{z}_0$, which is then decoded to obtain the generated image.

\begin{figure*}[!t]
    \centering
    \includegraphics[width=1.0\linewidth]{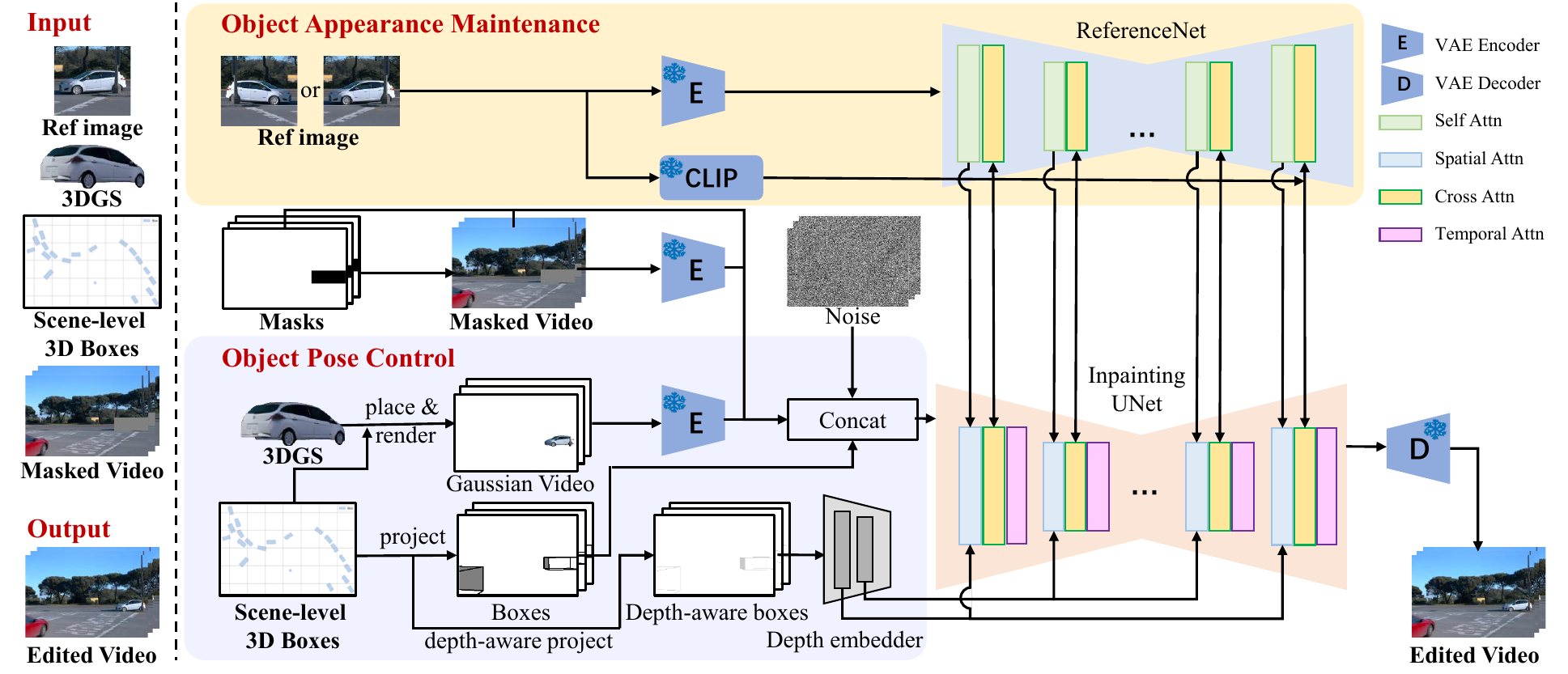}
    \caption{The overview of G\textsuperscript{2}Editor. A diffusion-based inpainting framework that includes object pose control and object appearance maintenance. This framework takes as input the assets of the edited object (the reference image and 3D Gaussian model), scene-level 3D boxes and the masked video, and outputs the edited video.}
    \label{fig:method}
\end{figure*}

Generally, image inpainting aims to fill masked regions of an image with reasonable content, enabling object insertion and deletion. Recently, some inpainting methods are built upon SD to enable photorealistic editing conditioned on text or reference images. Object editing in driving videos, which similarly involves repositioning, insertion and deletion, can be approximated as video inpainting. These inpainting UNets take as input the concatenation of the latent noise $\mathbf{z}_t$, the Variational Auto-Encoders (VAE) features of the background video with grayed-out foreground pixels $E(\mathbf{V}_{bg})$, and masks $\mathbf{M}$. $E$ is the VAE encoder. However, in driving videos, object editing emphasizes the control and coherence of the object pose, as well as the maintenance of appearance details. 
Therefore, we formulate object editing task based on inpainting task:
\begin{equation}
\label{eq:loss}
    \mathcal{L} = \mathbb{E}_{\mathbf{z}_t,\mathbf{c}_p,\mathbf{c}_a,\epsilon,t}(\left \| \epsilon -  \epsilon_{\theta}(\mathbf{z}_t,\mathbf{c}_p,\mathbf{c}_a,t) \right \| ^{2}_2),
\end{equation}
where $\mathbf{c}_p$ denotes pose-related conditions and $\mathbf{c}_a$ denotes appearance-related ones. 
Existing methods leverage 3D bounding boxes of the edited object for pose control and object image assets for appearance maintenance, achieving some success but still facing limitations. To enable more flexible and precise editing, we propose a pose control strategy that incorporates information from a 3D Gaussian model and scene-level 3D boxes (Sec.~\ref{sec:pose}), and maintain appearance using reference image features with random flipping (Sec.~\ref{sec:app}). Our inpainting framework is shown in Fig.~\ref{fig:method}.

\subsection{Object Pose Control}\label{sec:pose}
To enable precise control over object position and orientation, we introduce scene-level 3D boxes and the 3D Gaussian model as pose-related conditions $\mathbf{c}_p$.

We employ depth-aware boxes, similar to \cite{DriveEditor}. To maintain the layout of non-target objects while precisely controlling the edited object's poses, we use scene-level 3D boxes instead of boxes of the single edited object in \cite{DriveEditor}. Specifically,  each face of boxes is processed separately by projecting corner depths onto the image plane and interpolating within the face to form depth-aware boxes $\mathbf{D}_{b}$. These are encoded via a ResNet-style depth embedder to extract multi-scale features $f_l$, which are injected into ResBlocks of self-attention layers through a zero-init fusion layer:
\begin{equation}
    \mathbf{\nu}_l = \mathbf{\nu}_l + \text{Fusionlayer}(f_l),
\label{eq:fuselayer}
\end{equation}
where $\mathbf{\nu}_l$ is the feature map of $l$-th block, and the fusion layer consists of layer normalization, SiLU activation, and convolution operation. We also project the scene-level 3D boxes onto the image plane as edge masks $\mathbf{M}_{b}$.

To provide dense and accurate positional cues, we introduce a 3D Gaussian model beyond 3D bounding boxes. Specifically, we start with the 3D Gaussian asset of the edited object, which can be obtained from multiview reconstruction or image-to-3D generation. Each Gaussian is defined by its mean $\bm{\mu}$, rotation $\mathbf{R}$, scale $\mathbf{S}$, opacity $\alpha$ and color $\mathbf{o}$ in a local coordinate. We use the object pose $\{\mathbf{W}, \mathbf{T}\}$ to transform 3D Gaussians from local to world coordinate:
\begin{align}
\bm{\mu}_w &= \mathbf{W}\bm{\mu}+\mathbf{T}, \\
\mathbf{R}_w &= \mathbf{W}^T\mathbf{R}.
\end{align} 
The 3D Gaussians are then projected onto the 2D image plane using camera extrinsic and intrinsic. The rendered image of the edited object is computed by blending $N$ ordered Gaussians through the $\alpha$-rendering technique \cite{3DGS}:
\begin{equation}
\label{eq:color}
\mathbf{I}(\mathbf{p})=\sum_{i=1}^N \mathbf{o}_i \alpha_i \prod_{j=1}^{i-1}\left(1-\alpha_j\right),
\end{equation}
where $\mathbf{p}$ is the pixel of the rendering image. 
All rendered images make up the object's Gaussian video $\mathbf{V}_{g}=\{\textbf{I}_0,...,\textbf{I}_t\}$.

Since the Gaussian video is spatially aligned with the edited video, we concatenate their latents. Along with latent noise $\mathbf{z}_t$, the latent of the masked video $E(\mathbf{V}_{bg})$ and masks $\mathbf{M}$ used in the inpainting task, we concatenate the latent of the Gaussian video $E(\mathbf{V_g})$ and edge masks $\mathbf{M}_{b}$, yielding input for the inpainting UNet.

Some methods \cite{GenMM,DriveEditor} directly feed reference images and target object poses to the diffusion process, requiring the model to correctly understand the 3D scene geometry and the viewpoint transformation, an inherently challenging task. In contrast, taking the renderings of the edited object as input, G\textsuperscript{2}Editor edits on slightly blurred Gaussian images, which is more akin to Gaussian image restoration \cite{ReconDreamer}. This significantly simplifies the diffusion model's training objective and enhances control over object poses. Moreover, the Gaussian model explicitly ensures the 3D spatial coherence.

In summary, our pose‑related conditions $\mathbf{c}_p$ in (\ref{eq:loss}) comprise Gaussian video $\mathbf{V}_{g}$, edge masks of scene-level boxes $\mathbf{M}_{b}$ and depth‑aware boxes $\mathbf{D}_{b}$.

\subsection{Object Appearance Maintenance}\label{sec:app}
Object editing requires the maintenance and consistency of appearance details. Many image-driven generation methods use the CLIP image encoder to extract features, which are injected into the diffusion process via cross-attention. However, as noted in \cite{animate}, CLIP effectively captures high-level information but lacks the ability to preserve fine-grained details. Following \cite{GenMM}, we complement CLIP features $\mathbf{c}_c$ with a ReferenceNet that shares the architecture of the original SD, leveraging its pre-trained capability for extracting image features. Since the reference image and latent noise are not spatially aligned and only parts of latent noise require information from the reference, naive injection strategies such as concatenation or addition are suboptimal. Therefore, we replace the self-attention layers in the inpainting UNet with spatial-attention layers to selectively attend to relevant features. Specifically, spatial attention is performed over the concatenation of feature maps ($\mathbf{\nu}$ and $\mathbf{c}_r$) from the inpainting UNet and ReferenceNet:
\begin{equation}
    \mathbf{\nu}= \mathbf{\nu} + \text{TS}(\text{SelfAttn}([\mathbf{\nu},\mathbf{c}_r])),
\end{equation}
where TS(·) is a token selection operation that only retains the features from the inpainting UNet. The reference feature $\mathbf{c}_r$ is replicated $t$ times along the temporal dimension and concatenated with $\mathbf{\nu}$ along the spatial dimension. Overall, the appearance-related condition $\mathbf{c}_a$ in (\ref{eq:loss}) comprises CLIP features $\mathbf{c}_c$ and ReferenceNet features $\mathbf{c}_r$ from the reference image.

Moreover, we observe that selecting a single image from the video clip as the reference image done in \cite{GenMM} may lead the diffusion model to learn pose from the ReferenceNet. As a result, during inference, the model tends to overfit to the pose in the reference image, limiting pose control despite appearance maintenance. Simple random horizontal flipping of the reference image can prevent the diffusion model from learning object pose information from ReferenceNet.

\subsection{Training Strategy}\label{sec:train}
\subsubsection{Data Preparation}
Object repositioning and insertion can be defined as reconstructing the masked object in the video given the object asset and 3D bounding boxes. To train our inpainting framework, we prepare the dataset. Specifically, we select $N$ frames as a video $\mathbf{V}$$\in$$\mathbb{R}^{N\times 3 \times H \times W}$, each satisfying the following requirements: The edited object appears in all frames with sufficient size, and fewer than two other objects are within 3m around (less occlusion). Each video has a set of 3D bounding boxes. To support object translation and rotation, the 3D bounding boxes are projected onto the image plane and enlarged appropriately to form masks $\mathbf{M}$$\in$$\mathbb{R}^{N\times H \times W}$, resulting in the masked video $\mathbf{V}_{bg}$. Reference images of a clip are obtained by cropping square regions from frames in which the edited object is fully visible. To enable object deletion and realistic inpainting of the background within the masked regions, random masks are applied to regions without objects in images and incorporated into the training data.
And we use white images as Gaussian renderings and the reference image for object deletion.

\subsubsection{Two Stage Training} 
Our framework adopts a two-stage training strategy. In the first stage, the model learns to paint the specified object within the masked region of a single image according to the pose and appearance conditions. The optimization objective is defined in (\ref{eq:loss}), where pose-related conditions include Gaussian video, edge masks and depth-aware boxes, and appearance-related conditions consist of ReferenceNet features and CLIP features. 
To prevent the model from directly learning pose or background cues from the reference image, we select the reference and edited frames from the same clip with the largest temporal distance. At inference, the reference image is from external sources, and may differ significantly in lighting from the edited video. Therefore, we apply brightness, contrast and saturation augmentations to the reference image, enabling the model to infer lighting conditions from the unmasked background. In the second stage, we aim to enhance the temporal consistency of the edited region and restore the occluded background. Following \cite{animate,GenMM}, we incorporate temporal-attention layers after the cross-attention layers in the inpainting UNet, which are self-attention along the temporal dimension. At this stage, we only train the temporal layers.

\section{Experiments}\label{sec:exps}
\subsection{Experimental Setup}\label{sec:exp_setup}
\subsubsection{Dataset}
We construct a training dataset based on the Waymo Open Dataset \cite{waymo}. This autonomous driving dataset comprises 798 training scenes, each providing surround-view images along with 3D bounding box annotations at 10Hz. We use images from the three front-view cameras and resize them to $640\times 960$. We set $N=10$. After the processing described in Sec.~\ref{sec:train}, 19477 video clips are obtained for training. We also use images whose object-free regions are randomly masked at the first stage training with a probability of $0.2$. 

\begin{figure*}[t]
    \centering
    \includegraphics[width=1.0\linewidth]{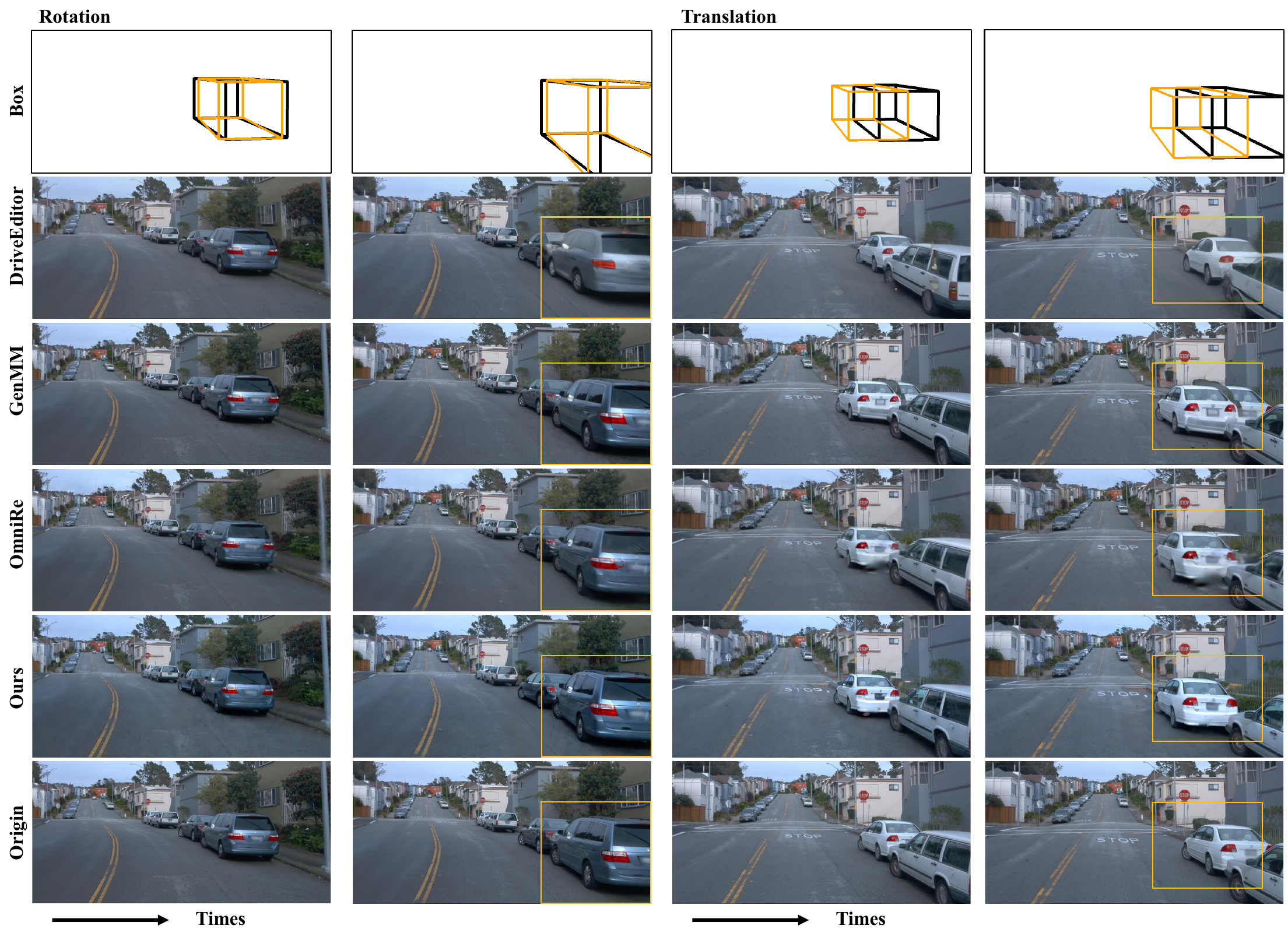}
    \caption{Visualization of rotation and translation. In the first row, black and orange boxes denote the object before and after manipulation. For rotation, G\textsuperscript{2}Editor achieves precise yaw control while maintaining appearance, outperforming 3DGS-based method and other video editing methods prone to artifacts or poor pose control. For translation, G\textsuperscript{2}Editor enables precise pose control and inpaints reasonable background, whereas other methods often leave residual artifacts.}
    \label{fig:exp1}
\end{figure*}

\subsubsection{Baselines}

Our work focuses on object editing in driving videos, emphasizing pose control and visual fidelity. There are limited comparable methods, and evaluation approaches also require further exploration. To evaluate fine-grained pose control and visual fidelity in edited regions, we design benchmarks for repositioning, insertion and deletion.

For repositioning, we apply the same data preparation to the Waymo validation set. For each target object, we perform three manipulations to assess fine-grained pose control: reinsertion (masking the object and inserting it at the same position), clockwise rotation by $\beta$ degrees and one-meter leftward translation. We compare our method with the 3DGS-based method OmniRe \cite{OmniRe}, as well as two driving video editing methods, GenMM \cite{GenMM} and DriveEditor \cite{DriveEditor}. For OmniRe, to enable editing of static vehicles, we model static and dynamic vehicles as rigid nodes. 
Note that the Gaussian videos used in our method are also derived from OmniRe. We use our own implementation of GenMM, aligned with our setup, and apply the official model for DriveEditor inference.

For insertion, we evaluate the visual fidelity, focusing on lighting and shadows. We use the SOTA image-to-3D method, TRELLIS \cite{TRELLIS}, to generate the 3D Gaussian model of the object from the reference image, scaling it to align with aggregated LiDAR points. These models are inserted into reconstructed scenes based on existing object trajectories, and the rendered video clips serve as a baseline. For our method, the Gaussian videos of the edited object provide guidance.

For deletion, we conduct a qualitative comparison with the 3DGS-based method.

\begin{table*}[t]
  \centering
  \scriptsize
  \caption{Quantitative results for pose control on object repositioning}
    \begin{tabular}{lccccccccc}
    \toprule
    & \multicolumn{3}{c}{reinsertion} & \multicolumn{3}{c}{rotation 5°} & \multicolumn{3}{c}{translation 1m} \\
\cmidrule(l{2pt}r{2pt}){2-4} \cmidrule(l{2pt}r{2pt}){5-7} \cmidrule(l{2pt}r{2pt}){8-10}     
& LET-mAP$\uparrow$ & LET-mAPH$\uparrow$ & LET-mAPL$\uparrow$ & LET-mAP$\uparrow$ & LET-mAPH$\uparrow$ & LET-mAPL$\uparrow$ & LET-mAP$\uparrow$ & LET-mAPH$\uparrow$ & LET-mAPL$\uparrow$ \\
    \midrule
    OmniRe\cite{OmniRe} & \underline{0.736} & \underline{0.727} & \underline{0.474} & 0.724 & \underline{0.714} &  \underline{0.478} & \underline{0.678} & \underline{0.668} & \underline{0.419} \\
    DriveEditor\cite{DriveEditor} & 0.562 & 0.546 & 0.318 & 0.532 & 0.511 & 0.306 & 0.280 & 0.266 & 0.152 \\
    GenMM\cite{GenMM} & 0.715 & 0.704 & 0.444 & \underline{0.726} & 0.709 & 0.440 & 0.416 & 0.408 & 0.217 \\
    Ours  & \textbf{0.781} & \textbf{0.772} & \textbf{0.504} & \textbf{0.806} & \textbf{0.794} & \textbf{0.517} & \textbf{0.725} & \textbf{0.715} & \textbf{0.463} \\
    \bottomrule
    \end{tabular}%
  \label{tab:exp1_pos}%
\end{table*}%

\begin{table}[t]
  \centering
  \scriptsize
  \caption{Quantitative results for appearance maintenance on object repositioning}
  \setlength{\tabcolsep}{0.005\linewidth}
    \begin{tabular}{lcccccc}
\toprule          & \multicolumn{2}{c}{reinsertion} & \multicolumn{2}{c}{rotation 5°} & \multicolumn{2}{c}{translation 1m} \\
\cmidrule(l{2pt}r{2pt}){2-3} \cmidrule(l{2pt}r{2pt}){4-5} \cmidrule(l{2pt}r{2pt}){6-7}        & LPIPS-Ref$\downarrow$ & FID-Ref$\downarrow$ & LPIPS-Ref$\downarrow$ & FID-Ref$\downarrow$ & LPIPS-Ref$\downarrow$ & FID-Ref$\downarrow$ \\
    \midrule
    OmniRe\cite{OmniRe} & 0.162 & 27.297 & 0.231 & 30.610 & 0.460 & 37.712 \\
    DriveEditor\cite{DriveEditor} & 0.274 & 25.994 & 0.276 & 25.659 & 0.516 & 33.435 \\
    GenMM\cite{GenMM} & \textbf{0.150} & \textbf{12.603} & \textbf{0.158} & \textbf{12.968} & \underline{0.456} & \underline{24.658} \\
    Ours  & \underline{0.151} & \underline{13.240} & \underline{0.207} & \underline{14.222} & \textbf{0.440} & \textbf{19.745} \\
    \bottomrule
    \end{tabular}%
  \label{tab:exp1_app}%
\end{table}%

\subsubsection{Metrics}
For repositioning, we evaluate both the pose control accuracy and appearance quality.
We select 8 challenging scenes from Waymo validation set, which contain editable objects and complex environmental conditions. 180 video clips are obtained for evaluation. We set $\beta = 5^\circ$. For object pose control accuracy, we apply the PGD model \cite{PGD}, a pre-trained vision-centric 3D object detector, to the edited videos. We use the LET metrics \cite{LET} (LET-mAP, LET-mAPH, LET-mAPL) provided by Waymo and set $5\%$ longitudinal error tolerance. As scene-level (full-image) metrics are not sensitive to edited object detection, we only consider the ground-truth/detection pairs of the edited instances. For appearance quality, we evaluate only the square region around the edited object to focus on the modifications. We resize the square images to $512\times 512$ and utilize frame-wise FID \cite{FID} and LPIPS \cite{lpips} between the edited and the original images.

For insertion, we evaluate visual fidelity. We use 3 Gaussian assets generated from TRELLIS and insert them into 9 scenes, resulting in 102 video clips. Since there is no ground-truth for the object's appearance in the target video, we consider the reference image
and the cropped square images. We also use FID and LPIPS.

\subsubsection{Training Setup}
We train the first stage for 30,000 iterations using a batch size of 8 per GPU on 4 A800 GPUs, and the second stage for 10,000 iterations using a batch size of 1 clip per GPU on 8 A800 GPUs. We initialize our inpainting UNet and ReferenceNet with SD v1.4-image-variations, and the extra 10 channels in the first layer of inpainting UNet are initialized to zero. We initialize the temporal attention layers with the pretrained motion module in \cite{AnimateDiff}.

\begin{table*}[htbp]
  \centering
  \scriptsize
  \caption{Ablation study on the repositioning task}
  \resizebox{\linewidth}{!}{
    \begin{tabular}{ccccccccccccccc}
    \toprule
      \multirow{3}{*}{\rotatebox{90}{\parbox{1.2cm}{3DGS}}}    & \multirow{3}{*}{\rotatebox{90}{\parbox{1.2cm}{Reference Net}}}      &  \multirow{3}{*}{\rotatebox{90}{\parbox{1.2cm}{random flip}}}      &  \multirow{3}{*}{\rotatebox{90}{\parbox{1.2cm}{scene-level boxes}}}     &  \multirow{3}{*}{\rotatebox{90}{\parbox{1.2cm}{temporal attn}}}     & \multicolumn{6}{c}{Object Pose Control}       & \multicolumn{4}{c}{Object Appearance Maintenance} \\
\cmidrule(l{2pt}r{2pt}){6-11} \cmidrule(l{2pt}r{2pt}){12-15}          &       &       &       &       & \multicolumn{3}{c}{rotation 5\degree} & \multicolumn{3}{c}{translation 1m} & \multicolumn{2}{c}{rotation 5\degree} & \multicolumn{2}{c}{translation 1m} \\
\cmidrule(l{2pt}r{2pt}){6-8} \cmidrule(l{2pt}r{2pt}){9-11} \cmidrule(l{2pt}r{2pt}){12-13} \cmidrule(l{2pt}r{2pt}){14-15}  & &  & &  & LET-mAP$\uparrow$ & LET-mAPH$\uparrow$ & LET-mAPL$\uparrow$ & LET-mAP$\uparrow$ & LET-mAPH$\uparrow$ & LET-mAPL$\uparrow$ & LPIPS-Ref$\downarrow$ & FID-Ref$\downarrow$ & LPIPS-Ref$\downarrow$ & FID-Ref$\downarrow$ \\
    \midrule
    \checkmark     & \checkmark     & \checkmark     & \checkmark     & \checkmark     & \textbf{0.806} & \textbf{0.794} & \textbf{0.517} & \textbf{0.725} & \textbf{0.715} & \textbf{0.463} & 0.207 & \textbf{14.222} & \textbf{0.440} & \textbf{19.745} \\
    $\checkmark$     & $\checkmark$     & $\checkmark$     & $\checkmark$     & $\times$     & 0.798 & 0.786 & 0.510 & 0.709 & 0.701 & 0.451 & 0.213 & 15.397 & 0.440 & 20.240 \\
    $\checkmark$     & $\checkmark$     & $\checkmark$     & $\times$     & $\times$     & 0.780 & 0.767 & 0.493 & 0.706 & 0.697 & 0.437 & 0.213 & 15.530 & 0.441 & 20.211 \\
    $\times$     & $\checkmark$     & $\checkmark$     & $\times$     & $\times$     & 0.668 & 0.649 & 0.396 & 0.416 & 0.407 & 0.212 & \textbf{0.201} & 14.289 & 0.471 & 19.804 \\
    $\checkmark$     & $\checkmark$     & $\times$     & $\times$     & $\times$     & 0.776 & 0.762 & 0.495 & 0.692 & 0.683 & 0.439 & 0.214 & 16.369 & 0.441 & 21.144 \\
    $\checkmark$     & $\times$     & $\times$     & $\times$     & $\times$     & 0.782 & 0.770 & 0.482 & 0.706 & 0.697 & 0.441 & 0.258 & 18.713 & 0.457 & 22.537 \\
    \bottomrule
    \end{tabular}%
    }
  \label{tab:abla}%
\end{table*}%

\begin{table*}[htbp]
  \centering
  \scriptsize
  \caption{Effect of edited data on detector performance under three-view camera setting}
  \resizebox{\linewidth}{!}{
    \begin{tabular}{lllcccccc}
    \toprule
    {Pretraining (25e)} & \multicolumn{2}{c}{Finetuning (+5e)} & \multicolumn{3}{c}{Front} & \multicolumn{3}{c}{Three Front Camera} \\
\cmidrule(l{2pt}r{2pt}){4-6} \cmidrule(l{2pt}r{2pt}){7-9}  Real data & Real data & Edited data & LET-mAP$\uparrow$ & LET-mAPH$\uparrow$ & LET-mAPL$\uparrow$ & LET-mAP$\uparrow$ & LET-mAPH$\uparrow$ & LET-mAPL$\uparrow$ \\
    \midrule
    $\checkmark$ & & & 0.612 & 0.606 & 0.453 & 0.536 & 0.529 & 0.395 \\
    $\checkmark$ & $\checkmark$ & & 0.624 & 0.618 & 0.463 & 0.554 & 0.547 & 0.409 \\
    $\checkmark$ & & $\checkmark$ & 0.628 & 0.622 & 0.466 & 0.556 & 0.548 & 0.411 \\
    \bottomrule
    \end{tabular}%
    }
    \label{tab:exp4_2}%
\end{table*}%

\subsection{Performance}
\subsubsection{Repositioning Objects}

Fig.~\ref{fig:exp1} shows qualitative editing results of $5^\circ$ rotation (left 2 columns) and $1m$ translation (right 2 columns) of the target object. For rotation, the 3DGS-based method can accurately control the rotation angle, but often produces blurring and artifacts due to overfitting to training views during Gaussian optimization. As shown in Fig.~\ref{fig:gs_effect}, this issue becomes increasingly evident as the rotation angle increases.
Other video editing methods struggle to respond to pose control inputs. For frames with large pose variations or those temporally distant from the reference image, DriveEditor fails to maintain object appearance. Our method enables precise control over the yaw of the edited object while maintaining high-fidelity appearance. 
For translation, DriveEditor fails to perform accurate movement, while GenMM and OmniRe can not remove objects from the original location. Our method enables precise control of object pose and inpaints reasonable background at the original location.

\begin{figure}[t]
    \centering
    \includegraphics[width=1.0\linewidth]{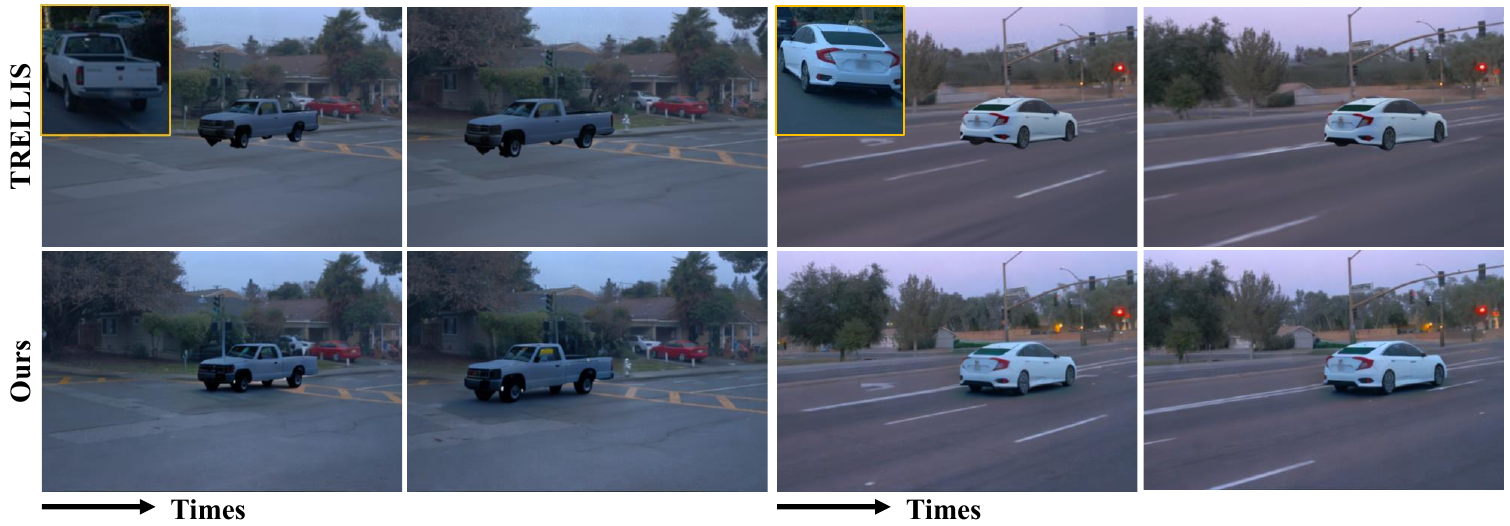}
    \caption{Based on image-to-3D generation method, G\textsuperscript{2}Editor is capable of synthesizing realistic shadows during object insertion.}
    \label{fig:exp2}
\end{figure}

\begin{figure}[t]
    \centering
    \includegraphics[width=1.0\linewidth]{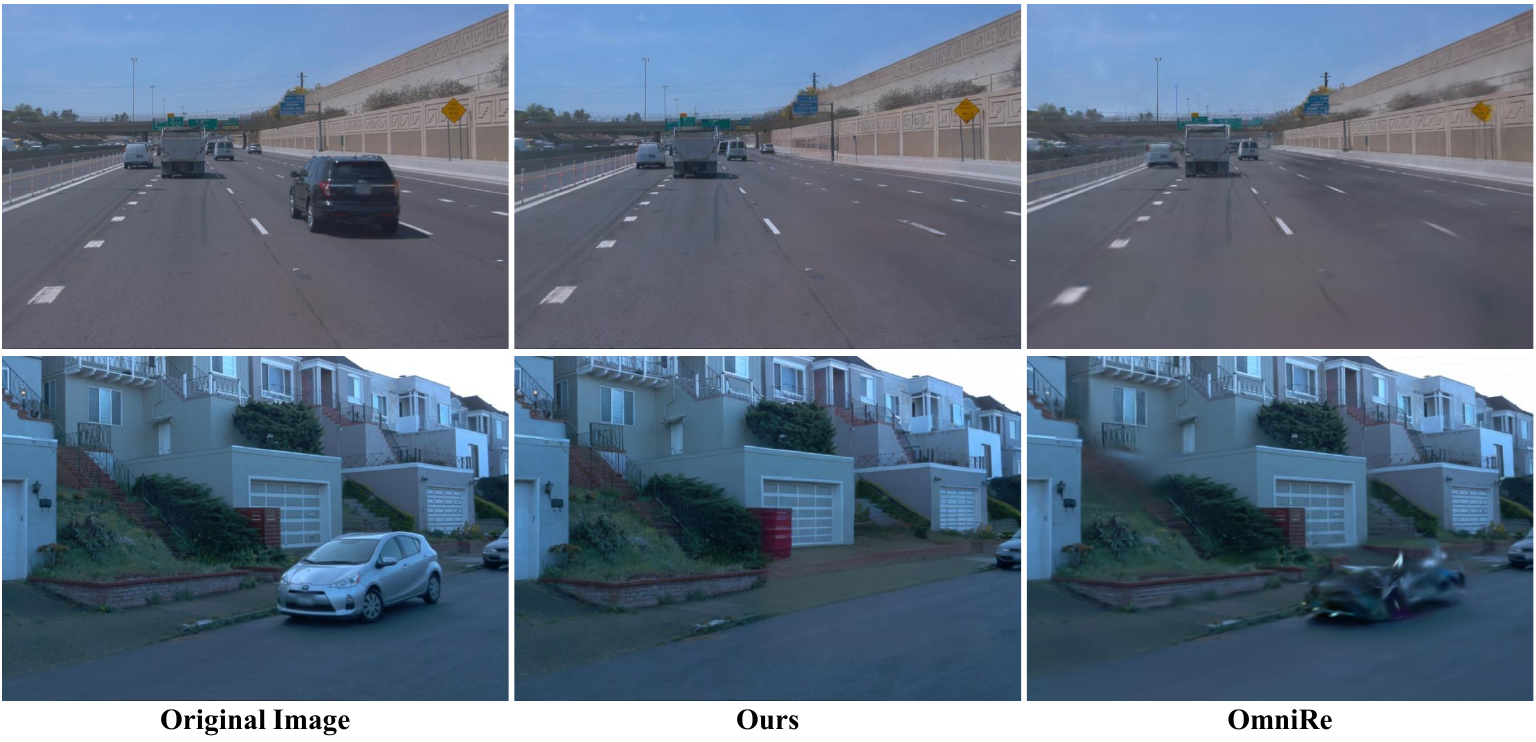}
    \caption{Visualization of object deletion. Compared to the 3DGS-based method, G\textsuperscript{2}Editor enables more realistic object deletion and reasonable background completion, particularly for static objects.}
    \label{fig:exp3}
\end{figure}

Table~\ref{tab:exp1_pos} and Table~\ref{tab:exp1_app} report the quantitative performance of our method and baselines on object pose control and appearance maintenance, respectively. In this experiment, our method significantly outperforms baselines on pose control over the edited objects. For appearance maintenance, our method greatly surpasses OmniRe and DriveEditor, and is on par with GenMM. We attribute this to the fact that GenMM generates parts of the images, which have higher resolution ($512\times 512$) for evaluated regions during inference.

\subsubsection{Inserting Objects}

Fig.~\ref{fig:exp2} shows qualitative results of object insertion using Gaussian models from TRELLIS directly and our Gaussian-guided method. Directly inserting 3D Gaussian models does not model the light source of the scene and cannot render realistic shadows. Our method can synthesize shadows and generate more realistic videos. 
For quantitative comparison, we measure the LPIPS-Ref metric for both TRELLIS and our method. Our method achieves 0.612 while TRELLIS is 0.639, showing better visual fidelity.

\subsubsection{Deleting Objects}
Fig.~\ref{fig:exp3} shows qualitative results of object deletion. 
For OmniRe, object deletion is achieved by removing the target’s Gaussian model when rendering, but this introduces undesirable background distortions (row 1). And the separation of static foreground and background makes static object deletion challenging (row 2). G\textsuperscript{2}Editor enables more realistic object deletion and reasonable background completion, particularly for static objects.

\begin{figure}[t]
    \centering
    \includegraphics[width=1.0\linewidth]{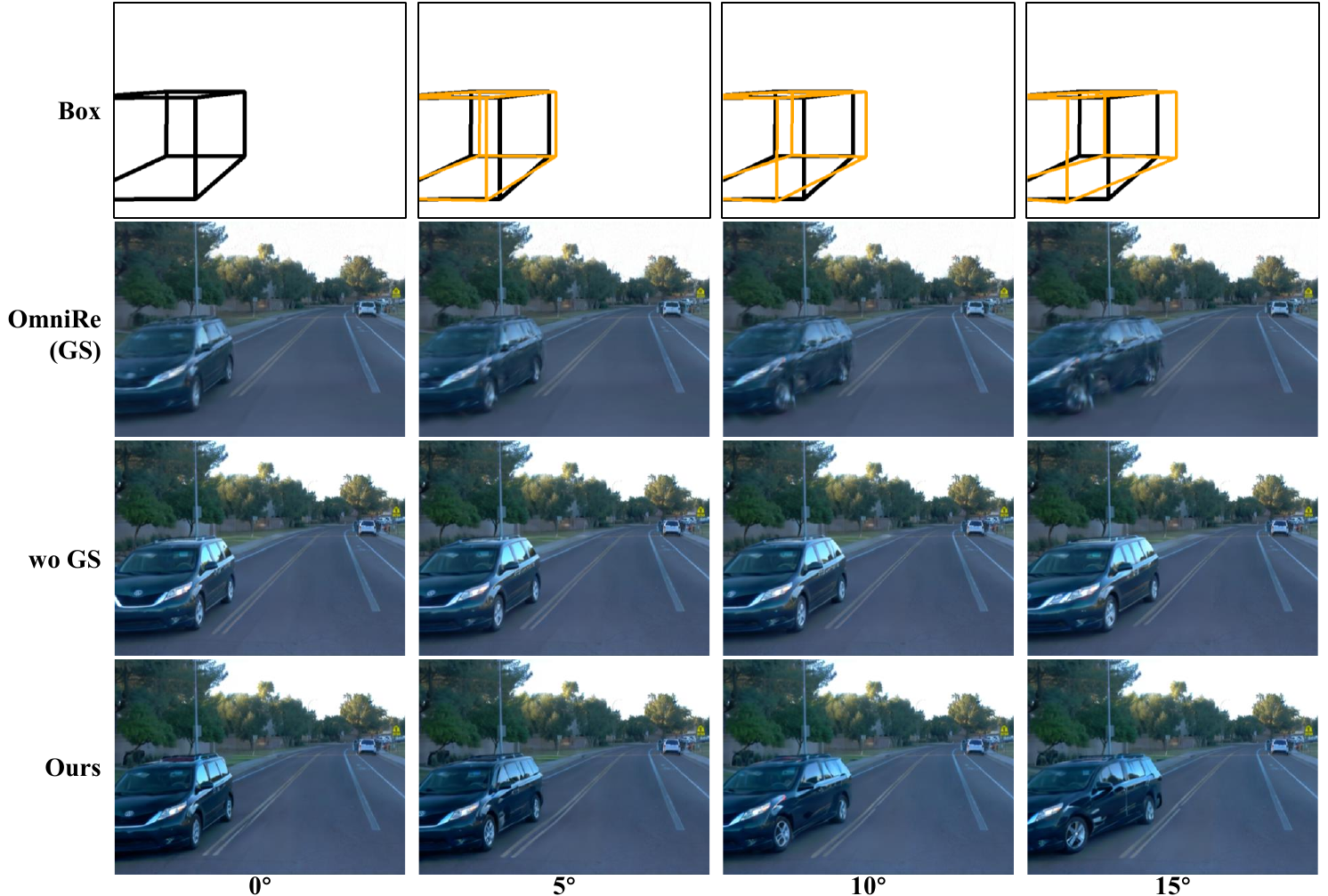}
    \caption{Effectiveness of 3DGS. By leveraging the dense and accurate spatial positional cues provided by 3DGS, our method achieves precise control over the target object's rotation, despite slightly blurred Gaussian renderings.}
    \label{fig:gs_effect}
\end{figure}

\begin{figure}[t]
    \centering
    \includegraphics[width=1.0\linewidth]{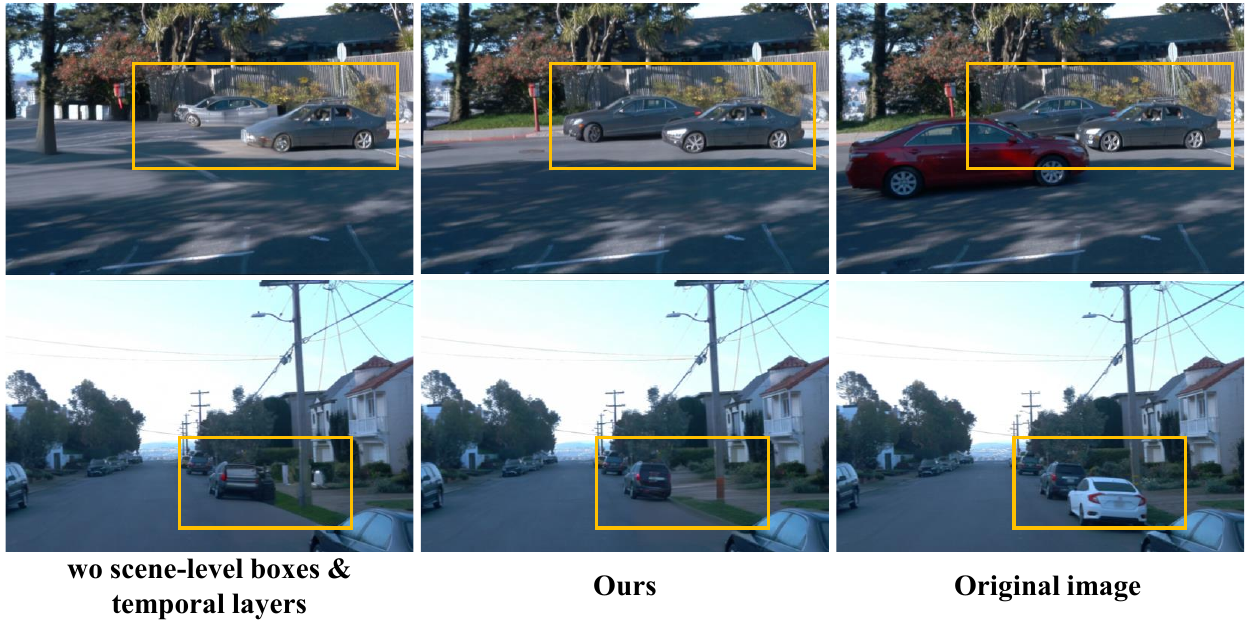}
    \caption{Effectiveness of scene-level boxes and temporal layers. Temporal layers and scene-level boxes contribute to the completion of occluded regions of non-target objects in each frame on object deletion.}
    \label{fig:box_effect}
\end{figure}

\subsection{Ablation Study}
To verify the effectiveness of each component, we conduct ablation studies on the repositioning task, evaluating both pose control and appearance maintenance. We first ablate the temporal layers to obtain an image editing framework. Based on this, we further ablate scene-level boxes, 3DGS renderings, the horizontal flipping of the reference image, and ReferenceNet. As shown in Table \ref{tab:abla}, the 3DGS rendering has a significant effect on the object pose control, whereas ReferenceNet is notably effective on the object appearance maintenance. Note that the model without the 3DGS rendering has better LPIPS when the edited objects are rotated by $5^\circ$. We attribute this to poor pose control. We find that the edited objects undergo minimal rotation in this setting, resulting in outputs more similar to original images. Additionally, random flipping the reference image horizontally indeed improves pose control accuracy by discouraging the framework from relying on ReferenceNet for object pose.

Fig.~\ref{fig:gs_effect} illustrates the effectiveness of 3DGS renderings in pose control. As shown in Fig.\ref{fig:box_effect}, scene-level boxes and temporal layers facilitate the completion of occluded regions of non-target objects in each frame on object deletion.

\subsection{Training Support for 3D Object Detection}
We generate edited data by repositioning (translating and rotating) objects in selected clips of Waymo training data, which is suitable for editing, aiming to enhance the training of the monocular object detector PGD\cite{PGD}. PGD is trained for 25 epochs following the official setting, and then fine-tuned for 5 additional epochs with real data combined with edited data on three front-view cameras. We copy the corresponding real data to ensure the same total data volume for fair comparison. We evaluate performance using the official Waymo scene-level LET metrics\cite{LET}. As shown in Table~\ref{tab:exp4_2}, repositioning effectively expands the viewpoint distribution of objects and increases data diversity, thereby improving detection performance compared to simple duplication of real data.
\section{Conclusion}\label{sec:conclu}

We propose G\textsuperscript{2}Editor, a diffusion-based framework that supports object repositioning, insertion and deletion in driving videos. To enable precise control over object pose, we introduce a hybrid control strategy combining 3DGS renderings and 3D boxes as pose-related conditions. And we hierarchically inject visual features from a random-flipped reference image into the diffusion process to maintain the edited object’s appearance. We design fine-grained experiments to evaluate the accuracy of pose control and visual fidelity within the edited region, demonstrating that G\textsuperscript{2}Editor outperforms existing methods and benefits downstream AD tasks.

\bibliographystyle{ieeetr}
\bibliography{ref}

\end{document}